\documentclass[lettersize,journal]{IEEEtran}
\usepackage{amsmath,amsfonts}
\usepackage{algorithmic}
\usepackage{array}
\usepackage[caption=false,font=normalsize,labelfont=sf,textfont=sf]{subfig}
\usepackage{textcomp}
\usepackage{stfloats}
\usepackage{url}
\usepackage{verbatim}
\usepackage{graphicx}
\usepackage{booktabs}
\usepackage{cite}

\def\BibTeX{{\rm B\kern-.05em{\sc i\kern-.025em b}\kern-.08em
    T\kern-.1667em\lower.7ex\hbox{E}\kern-.125emX}}
\usepackage{balance}
\usepackage{multirow}
\usepackage{amssymb,bbm}
\usepackage[ruled,linesnumbered]{algorithm2e}
\usepackage{bm}

\usepackage{CJKutf8}
\usepackage{amsmath}
\usepackage{pifont}

\DeclareMathOperator*{\argmax}{arg\,max}

\begin{document}
\title{Enhancing the Robustness of Contextual ASR to Varying Biasing Information Volumes Through Purified Semantic Correlation Joint Modeling}
\author{Yue Gu, Zhihao Du, Ying Shi, Shiliang Zhang, Qian Chen, and Jiqing Han, \textit{Member, IEEE}
\thanks{Yue Gu and Jiqing Han are with the Research Center of Auditory Intelligence, the School of Computer Science and Technology, the Faculty of Computing, Harbin Institute of Technology (e-mail: 427gy@sina.com, jqhan@hit.edu.cn).}}

\markboth{Journal of \LaTeX\ Class Files,~Vol.~18, No.~9, September~2020}%
{How to Use the IEEEtran \LaTeX \ Templates}

\maketitle
\begin{abstract}
  Recently, cross-attention-based contextual automatic speech recognition (ASR) models have made notable advancements in recognizing personalized biasing phrases. However, the effectiveness of cross-attention is affected by variations in biasing information volume, especially when the length of the biasing list increases significantly. We find that, regardless of the length of the biasing list, only a limited amount of biasing information is most relevant to a specific ASR intermediate representation.
  Therefore, by identifying and integrating the most relevant biasing information rather than the entire biasing list, we can alleviate the effects of variations in biasing information volume for contextual ASR.
  To this end, we propose a purified semantic correlation joint modeling (PSC-Joint) approach. In PSC-Joint, we define and calculate three semantic correlations between the ASR intermediate representations and biasing information from coarse to fine: list-level, phrase-level, and token-level. Then, the three correlations are jointly modeled to produce their intersection, so that the most relevant biasing information across various granularities is highlighted and integrated for contextual recognition. In addition, to reduce the computational cost introduced by the joint modeling of three semantic correlations, we also propose a purification mechanism based on a grouped-and-competitive strategy to filter out irrelevant biasing phrases. Compared with baselines, our PSC-Joint approach achieves average relative F1 score improvements of up to 21.34\% on AISHELL-1 and 28.46\% on KeSpeech, across biasing lists of varying lengths.
\end{abstract}

\section{Introduction}
\IEEEPARstart{I}{n} recent years, remarkable advancements have been made on end-to-end automatic speech recognition (E2E ASR), such as connectionist temporal classification \cite{DBLP:conf/icml/GravesFGS06}, recurrent neural network transducer \cite{DBLP:conf/icassp/GravesMH13,DBLP:conf/icml/GravesJ14}, and attention encoder-decoder \cite{DBLP:conf/nips/ChorowskiBSCB15,DBLP:conf/icassp/ChanJLV16,DBLP:conf/nips/VaswaniSPUJGKP17,DBLP:conf/interspeech/GulatiQCPZYHWZW20}. E2E ASR models have outperformed traditional hybrid speech recognition systems \cite{DBLP:journals/taslp/PrabhavalkarHSSW24}. 
Despite these advancements, the performance of E2E ASR models degrades significantly when recognizing personalized contexts such as the names of contacts, songs, and locations. To alleviate the degradation, contextual ASR has been introduced, which incorporates flexible context and improves the recognition performance on biasing phrases. In contextual ASR, personalized phrases are typically considered as biasing information and are introduced into the model as a biasing list.

Contextual speech recognition approaches can be broadly divided into four categories: shallow fusion approaches, prompt-based contextual approaches, trie-based deep biasing techniques, and attention-based deep context approaches.
Shallow fusion approaches \cite{DBLP:conf/interspeech/AleksicGMAHRRM15,hall15_interspeech,DBLP:conf/interspeech/WilliamsKARS18,DBLP:conf/icassp/ChenJWSF19,DBLP:conf/interspeech/ZhaoSRRBLP19,DBLP:conf/icassp/HeSPMAZRKWPLBSL19,huang20f_interspeech,DBLP:conf/interspeech/Meng0K0CYSL021,DBLP:conf/interspeech/NigmatulinaMVMZ23,10447571} integrate biasing information through a weighted finite state transducer or language model during the inference stage, going against the benefits derived from the joint optimization in sequence-to-sequence models. Prompt-based contextual approaches leverage the acoustic-textual alignment capability of large-scale multimodal speech models by treating the biasing phrase as the prompt, improving the recognition of biasing phrases\cite{yang2024ctc,yang2024mala}. Trie-based deep biasing approaches \cite{DBLP:conf/interspeech/LeJKKSMCSFKSS21,DBLP:conf/slt/LeKCMFS21,DBLP:conf/asru/SunZW21,DBLP:conf/interspeech/HardingTW23,DBLP:journals/taslp/SunZW23,10447652} focus on matching the prefix tree of biasing phrases using recognized partial tokens, where only some word pieces of the biasing phrase are considered. In contrast, attention-based deep context approaches \cite{DBLP:conf/slt/PundakSPKZ18, DBLP:conf/icassp/0002PS19, DBLP:conf/interspeech/JainKMZMS20,DBLP:conf/icassp/HanDZX21,DBLP:conf/icassp/LiuLZ21,DBLP:conf/asru/ChangLRMORK21,DBLP:conf/icassp/HanDLCZMX22,DBLP:conf/icassp/MunkhdalaiSCGCS22,DBLP:conf/interspeech/HuangZYGMX023,DBLP:conf/interspeech/BleekerSBZ23,DBLP:conf/interspeech/YangSWZM023,DBLP:conf/interspeech/NaowaratHDTH23,DBLP:conf/icassp/XuLHSWKM23,10096677,DBLP:conf/asru/JalalPPMSKZDLLJ23,DBLP:conf/interspeech/XuYHGZLCL023,DBLP:conf/icassp/AlexandridisSSCRSM23,DBLP:conf/icassp/FuSGLSMM23,wu23e_interspeech,sudo2024contextualized,10447438,10448264,10446106} encode all word pieces or tokens of a biasing phrase into an embedding vector. In such approaches, the cross-attention module uses ASR intermediate representations as queries to attend to all biasing phrases simultaneously. Due to their simplicity and flexibility, cross-attention-based contextual ASR models have attracted significant attention.

In cross-attention-based approaches, the correlations between ASR intermediate representations and biasing phrases are learned by the cross-attention module, which identifies the relevant biasing information from the given biasing list. 
However, in real-world applications, the effectiveness of the attention module is affected by variations of biasing information volumes \cite{10446106,DBLP:conf/icassp/HanDLCZMX22}, particularly when the number of biasing phrases is large, introducing irrelevant biasing information to the contextual ASR.
Although injecting distractors into the biasing list during training allows the cross-attention module to handle a longer biasing list, it also introduces more confusion between the target phrase and its similar ones when the number of injected distractors is large \cite{DBLP:conf/slt/PundakSPKZ18, DBLP:conf/icassp/HanDLCZMX22}. Furthermore, injecting more distractors improves performance on larger biasing lists but degrades performance on smaller lists \cite{DBLP:conf/interspeech/LeJKKSMCSFKSS21}, indicating that it cannot enhance the robustness of contextual ASR models to variations in biasing list length.



We find that, no matter how long the biasing list is, only a limited amount of biasing information is relevant to a specific ASR intermediate representation.
Therefore, we can mitigate the impact of biasing information volume variations for contextual ASR by identifying and integrating the most relevant biasing information rather than the entire biasing list.
To this end, we propose a novel purified semantic correlation joint modeling (PSC-Joint) approach, in which three semantic correlations are defined and calculated from coarse to fine: list-level, phrase-level, and token-level.
Then, based on the cross-granularity consistency of the most relevant biasing information, the three types of semantic correlations are jointly modeled to highlight and integrate this information through their intersection.
In addition, during the inference stage, a divide-and-conquer-based purification mechanism is also proposed to filter out the irrelevant phrases and refine the biasing list, thereby enhancing the computational efficiency of the multi-level semantic correlation joint modeling. 
\begin{figure}[t!]
	\centering
	\includegraphics[width=\linewidth]{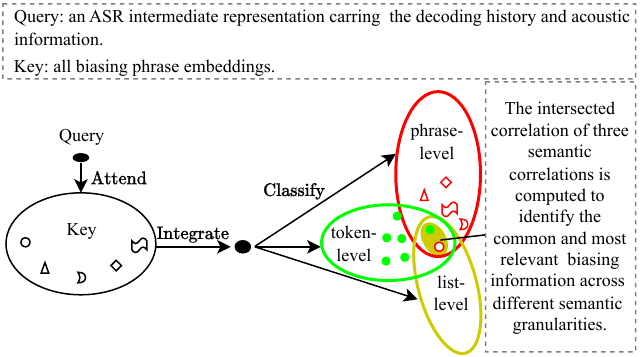}
	\caption{The schematic diagram of semantic correlation joint modeling method.}
	\label{fig:motivation}
\end{figure}

Fig.~\ref{fig:motivation} shows more details about the semantic correlation joint modeling method, in which an ASR intermediate representation and all biasing phrases serve as ``Query'' and ``Key'', respectively. 
For the list-level correlation, the biasing list is treated as a single entity, and the correlations between ``Query'' and ``Key'' are modeled as a binary prediction task, representing the correlation score of the ``Query'' being relevant to the entire biasing list. Similarly, the phrase-level correlation is modeled as a phrase classification task by treating each biasing phrase as an independent entity, representing the correlation score of the ``Query'' belonging to a specific biasing-phrase category. For the token-level correlation, each token in the biasing list is considered as an entity. The correlations are modeled as a token classification task, representing the correlation score of the ``Query'' belonging to a specific ASR vocabulary token.  
Obviously, the most relevant biasing information are consistent across three types of semantic correlations when the ``Query'' is relevant to any ``Key'' and it can be identified by the intersected correlations.
Thus, we compute the joint score of three semantic correlations and the containment relationship between phrases and tokens to produce the intersected correlations, in which the most relevant information is spotlighted. As a result, the proposed semantic correlation joint modeling method can significantly enhance the robustness of contextual ASR against changes in the biasing information volume.

To reduce the computational cost introduced by the joint modeling of three semantic correlations, we further propose a divide-and-conquer-based biasing information purification mechanism, which purifies biasing phrases in a grouped and competitive manner before semantic correlation joint modeling. 
Experiments conducted on the AISHELL-1, KeSpeech, and AISHELL-NER datasets show that PSC-Joint significantly improves the F1 score and reduces character error rate across varying-length biasing lists, outperforming the baseline models.

\section{RELATED WORK}
\noindent Various attention-based deep context approaches for autoregressive (AR) ASR and non-autoregressive (NAR) ASR \cite{DBLP:conf/interspeech/Higuchi0COK20, DBLP:conf/icassp/Dong020,DBLP:journals/spl/ChenWVZD21,DBLP:conf/icassp/Song0HW0M21,DBLP:conf/interspeech/NozakiK21} are developed recently. During inference, the AR decoder generates tokens one by one since each token is conditioned on all previous tokens. In contrast, the NAR decoder generates output sequences iteratively or simultaneously by removing temporal dependency \cite{DBLP:conf/interspeech/GaoZ0Y22}. The primary distinction between AR and NAR decoding is the time consumption of token generation, thus most attention-based approaches can be expanded for both AR and NAR ASR models in theory. The continuous Integrate-and-Fire (CIF) \cite{DBLP:conf/icassp/Dong020} based NAR ASR model is often employed as the ASR backbone considering its faster inference speed and comparable recognition performance. 

Recently, several attention-based deep context algorithms have been developed for CIF-based ASR. Within the CIF-based collaborative decoding (ColDec) framework for contextual ASR \cite{DBLP:conf/icassp/HanDZX21}, an attention-based context processing network (CPN) is employed to produce the context detection results, in which the CIF outputs are used as the queries and the biasing phrases are regarded as the keys and values. The target of CPN consists only of the biasing phrase and the ``\textless no-bias\textgreater'' token. For instance, the target of the CPN is set to ``\# peace\_ dove\_ \# \#'', whereas the ASR backbone target is ``the\_ peace\_ dove\_ symbolizes\_ peace\_'', where ``peace\_ dove\_'' represent a biasing phrase and ``\#'' denotes the special ``\textless no-bias\textgreater'' token.
At each output step in inference, the ASR decoder and CPN decoder conduct collaborative decoding with interpolated log-probability. 
To alleviate the token-level prediction uncertainty caused by the confusion between similar context-specific phrases, the fine-grained contextual knowledge selection (FineCoS) method employs a phrase selection (PS) strategy to narrow down phrase candidates and applies token-level attention to the selected phrases \cite{DBLP:conf/icassp/HanDLCZMX22}.
FineCoS absorbs the thought of collaborative decoding and further improves the ability to recognize the given biasing phrases.  
Similar to FineCoS, semantic-augmented contextual paraformer (SeACo) \cite{10446106} also employs the cross-attention module and the collaborative decoding method to produce the contextualized results. In SeACo, not only the CIF outputs but also the ASR backbone recognition outputs are fed into the cross-attention module in which the context detection outputs of the cross-attention module are semantic-augmented. SeACo utilizes an attention score filtering (ASF) strategy to discard irrelevant biasing phrases, by summing the attention scores among the sentence for each biasing phrase and picking the most active candidates based on these scores for the second-pass decoding. 
Notably, the PS and ASF strategies filter out the irrelevant biasing phrases according to the attention score, which may demonstrate limited effectiveness as the length of biasing lists increases significantly. 






\section{PURIFIED SEMANTIC CORRELATION JOINT MODELING APPROACH}
\subsection{Overview and Symbol Definition}
\label{chap:problem}
\begin{figure}[t!]
	\centering
	\includegraphics[width=\linewidth]{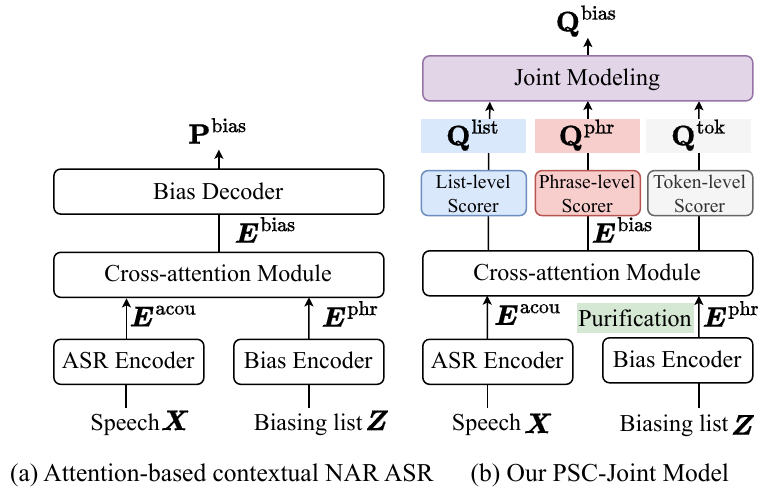}
	\caption{Schematic diagram of (a) current attention-based contextual NAR ASR model and (b) our PSC-joint Model.}
	\label{fig:sec3a}
\end{figure} 
\noindent Fig.~\ref{fig:sec3a}(a) illustrates that attention-based contextual NAR ASR models, such as ColDec \cite{DBLP:conf/icassp/HanDZX21}, typically consist of four components to generate transcriptions biased toward the provided phrases. 
First, an ASR encoder encodes a sequence of $T$ speech features $\boldsymbol{X}=\{\boldsymbol{x}_t \mid t=1,\cdots,T\}$ into acoustic hidden representations $\boldsymbol{E}^{\mathrm{acou}}=\{\boldsymbol{e}^{\mathrm{acou}}_u \mid u=1,\cdots,U\}$. Second, a biasing encoder transforms the biasing phrase list $\boldsymbol{Z}=\{\boldsymbol{z}_m \mid m=1,\cdots,M\}$ into phrase embeddings $\boldsymbol{E}^{\mathrm{phr}}=\{\boldsymbol{e}^{\mathrm{phr}}_m \mid m=1,\cdots,M\}$. $U$ and $M$ denote the transcription length of an utterance and the length of the biasing list, respectively. Third, a cross-attention module integrates all biasing phrases $\boldsymbol{E}^{\mathrm{phr}}$ to produce biased hidden representations $\boldsymbol{E}^{\mathrm{bias}}=\{\boldsymbol{e}^{\mathrm{bias}}_u \mid u=1,\cdots,U\}$, according to the correlations between $\boldsymbol{E}^{\mathrm{acou}}$ and $\boldsymbol{E}^{\mathrm{phr}}$. Finally, a bias decoder transform the biased hidden representations $\boldsymbol{E}^{\mathrm{bias}}$ into the token-level probability distributions $\mathbf{P}^{\mathrm{bias}}=\{\mathbf{P}^{\mathrm{bias}}_u  \in \mathbb{R}^{V} \mid u=1,\cdots,U\}$ over a vocabulary of $V$ tokens, yielding the biased output. Additionally, a vanilla ASR model is also involved as backbone, whose decoder transforms $\boldsymbol{E}^{\mathrm{acou}}$ into token probabilities $\mathbf{P}^{\mathrm{bb}}=\{\mathbf{P}^{\mathrm{bb}}_u  \in \mathbb{R}^{V} \mid u=1,\cdots,U\}$ over the same vocabulary. Then, $\mathbf{P}^{\mathrm{bb}}$ is interpolated with biased output $\mathbf{P}^{\mathrm{bias}}$ to yield the final recognition results.

At each decoding step $u$, the correlation $\boldsymbol{corr}_u$ between the acoustic hidden representation $\boldsymbol{e}^{\mathrm{acou}}_u$ and the set of phrase embeddings $\boldsymbol{E}^{\mathrm{phr}}$ is quantified by the attention scores that are typically computed using a scaled dot-product attention mechanism, formulated as $\mathit{corr}_{u,m}=\frac{1}{\sqrt{d}}<\boldsymbol{e}^{\mathrm{acou}}_u,\boldsymbol{e}^{\mathrm{phr}}_m>$. This correlation is then used to perform a weighted sum over embeddings in $\boldsymbol{E}^{\mathrm{phr}}$, resulting in a biased representation $\boldsymbol{e}_u^{\mathrm{bias}}$. 
\begin{align}
  \boldsymbol{e}_u^{\mathrm{bias}}&=\sum_{m=1}^{M}{corr_{u,m}\boldsymbol{e}^{\mathrm{phr}}_m}
  \end{align}
Subsequently, $\boldsymbol{e}_u^{\mathrm{bias}}$ is  passed through the bias decoder to produce probabilities $\mathbf{P}^{\mathrm{bias}}_u$ of the biased output.
Since the acoustic hidden representation $\boldsymbol{e}^{\mathrm{acou}}_u$ can be mapped to $V$ vocabulary tokens, $\boldsymbol{corr}_u$ reflects the correlations between the $M$ biasing phrases and the $V$ vocabulary tokens, residing in a space of $\mathbb{R}^{M\times V}$. However, during inference, the dimension of the correlation space can fluctuate significantly with the varying list length $M$, which may impair the integration of biasing phrases (Eq.~(1)) and introduce irrelevant biasing information into the contextual ASR model.

 By defining semantic correlations at multiple granularities and computing their intersection, the most relevant biasing information can be effectively identified. As this information is inherently limited regardless of the list length, leveraging it helps mitigate the impact of list length variations in contextual ASR models.
 Thus, we propose a semantic correlation joint modeling (SC-Joint) method, as illustrated in Fig.\ref{fig:sec3a}(b). Specifically, three semantic correlation scores at various granularities, i.e., list-level $\mathbf{Q}^{\mathrm{list}} = \{\mathrm{Q}^{\mathrm{list}}_u\in[0,1] \mid u=1,\cdots,U\}$, phrase-level $\mathbf{Q}^{\mathrm{phr}} = \{\mathbf{Q}^{\mathrm{phr}}_u\in\mathbb{R}^{M} \mid u=1,\cdots,U\}$, and token-level $\mathbf{Q}^{\mathrm{tok}} = \{\mathbf{Q}^{\mathrm{tok}}_u\in\mathbb{R}^{V} \mid u=1,\cdots,U\}$, are produced by three scorers, respectively. These scorers are trained by list-level, phrase-level, and token-level classification losses, respectively. Then, these correlations are jointly modeled by computing their intersection $\mathbf{Q}^{\mathrm{bias}} = \{ \mathbf{Q}^{\mathrm{bias}}_u \in \mathbb{R}^{{V}} \mid u=1,\cdots,U\}$, in order to identify the most relevant biasing information consistent across three levels:
 \begin{align}
  \mathbf{Q}^{\mathrm{bias}}_u = \mathbb{N}( \mathrm{Q}^{\mathrm{list}}_u (\mathbf{Q}^{\mathrm{phr}}_u)^\top \boldsymbol{\Phi} \mathbf{Q}^{\mathrm{tok}}_u ) \label{eq:jointmodel}
\end{align}
where $\mathbb{N}$ denotes a normalization function that ensures $V$ sub-scores are positive and sum to 1, such as the softmax function.
Note that the joint modeling takes into account the inclusion relationship between $V$ ASR vocabulary tokens and $M$ biasing phrases, represented by $\boldsymbol{\Phi}\in\mathbb{R}^{M\times V}$.

To reduce the computational cost of SC-Joint, we further propose a divide-and-conquer-based biasing information purification mechanism, which is performed before SC-Joint, composing the PSC-Joint approach. In each round of purification, the biasing phrase embedding $\boldsymbol{E}^{\mathrm{phr}}$ is divided into $G$ groups, then the irrelevant phrases in each group $\boldsymbol{E}^{\mathrm{phr}}_{g}$ are competitively filtered out in parallel based on the list-level $\mathbf{Q}_{g}^{\mathrm{list}}$ and phrase-level $\mathbf{Q}_{g}^{\mathrm{phr}}$ correlations:
\begin{align}
  \boldsymbol{E}^{\mathrm{phr}} \in \mathbb{R}^{M}\!\xrightarrow{\text{Group}}\!\{\boldsymbol{E}^{\mathrm{phr}}_{g} \in \mathbb{R}^{group\_size} \mid g=1,\cdots,G\}
\end{align}

\begin{equation}
  \begin{aligned}
    \left\{
      \begin{aligned}
      \left(\boldsymbol{E}^{\mathrm{phr}}_{g},\boldsymbol{E}^{\mathrm{acou}}\right)
      \xrightarrow{\text{Attend}} \boldsymbol{E}^{\mathrm{bias}}_g  \xrightarrow{\text{Score}} \left(\mathbf{Q}^{\mathrm{list}}_g, \mathbf{Q}^{\mathrm{phr}}_g\right)
      \end{aligned}
      \right\}_{g=1}^{G}\\
       \xrightarrow{\text{Topk (group-wise) and Merge}} \boldsymbol{E}^{\mathrm{phr}} \in \mathrm{R}^{M^{\mathrm{pur}} }
  \end{aligned}
\end{equation}

where $M^{\mathrm{pur}}$ represents the number of retained biasing phrases with $M^{\mathrm{pur}} \ll M$. In this way, many irrelevant phrases are discarded. Then, the multi-level semantic correlations can be predicted more accurately on the shorter biasing list.
In addition to discarding irrelevant phrases, the purification mechanism also reduces the number of biased vocabulary tokens by eliminating tokens not included in the refined list. Thus, this mechanism enhances both computational efficiency and the identification of the most relevant biasing information.

Finally, the recognition results of ASR backbone $\mathbf{P}^{\mathrm{bb}}$ can be interpolated with the joint modeling results $\mathbf{Q}^{\mathrm{bias}}$ automatically according to the list-level correlation $\mathbf{Q}^{\mathrm{list}}$ to obtain the contextualized results $\mathbf{Q}^{\mathrm{casr}}=\{\mathbf{Q}^{\mathrm{casr}}_u \in \mathbb{R}^V \mid u=1,\cdots,U\}$:
\begin{equation}
  \mathbf{Q}^{\mathrm{casr}} = (\mathrm{1}-\mathbf{Q}^{\mathrm{list}})\mathbf{P}^{\mathrm{bb}} + \mathbf{Q}^{\mathrm{list}}\mathbf{Q}^{\mathrm{bias}}
  \label{eq:coldec}
\end{equation}
\begin{figure*}[t!]
	\centering
	\includegraphics[width=\linewidth]{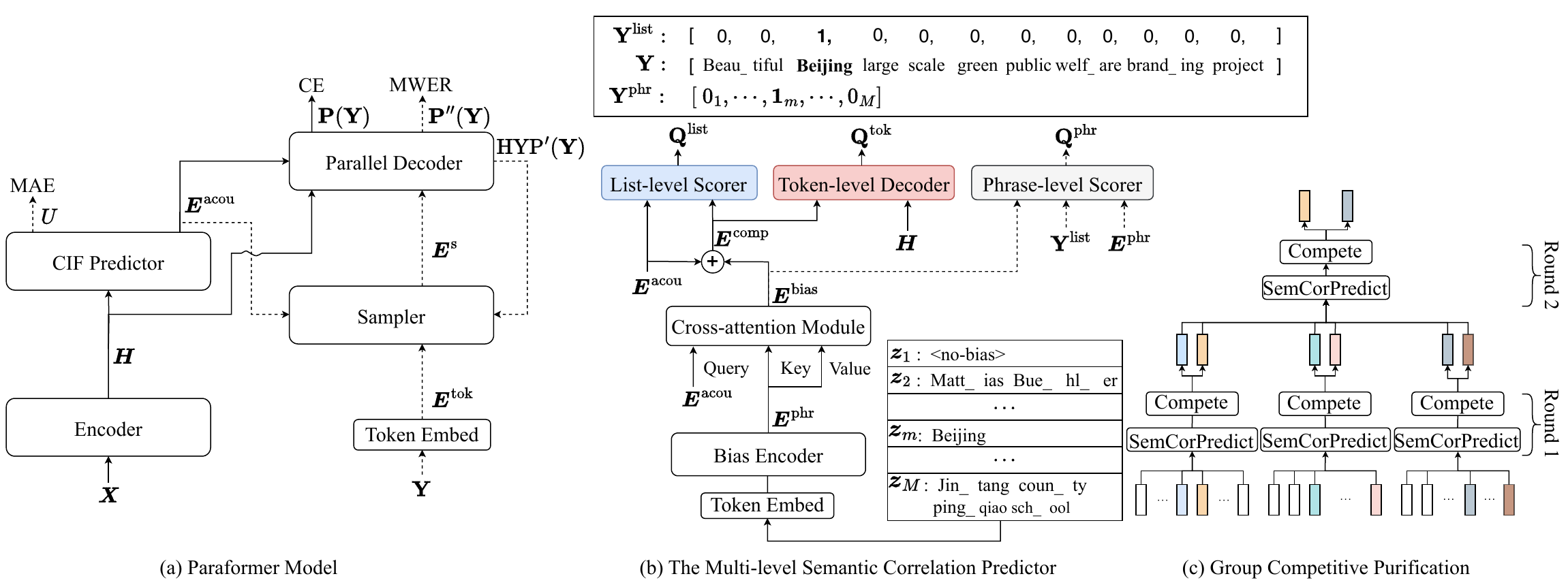}
	\caption{Schematic diagram of the purified semantic correlation joint modeling approach. (a) Non-autoregressive ASR backbone model: Paraformer. (b) Multi-level semantic correlation predictor (SemCorPredict). (c) Purification flowchart. In Figures (a) and (b), solid lines denote inference paths, whereas dotted lines represent connections used exclusively during training.}
	\label{fig:overview}
\end{figure*} 
\subsection{ASR Backbone}
\label{chap:backbone}
The powerful NAR ASR model, Paraformer \cite{DBLP:conf/interspeech/GaoZ0Y22}, is the backbone to obtain the basic recognition results.
As shown in Fig.~\ref{fig:overview}(a), the Paraformer consists of an encoder, a continuous integrate-and-fire (CIF) predictor, and a parallel decoder.
Firstly, the speech feature $\boldsymbol{X}$ is fed into the encoder to extract the speech representation $\boldsymbol{H}=\{\boldsymbol{h}_t \mid t=1,\cdots,T\}$.
Then, the CIF predictor accumulates $\boldsymbol{H}$ by weighted summation and predicts the acoustic embeddings $\boldsymbol{E}^{\mathrm{acou}}$ for each ASR output token.
In the parallel decoder, $\boldsymbol{E}^{\mathrm{acou}}$ integrates the acoustic information of $\boldsymbol{H}$ to predict the probabilities of a sequence of target tokens, $\mathbf{Y}=\{\mathrm{y}_u \mid u=1,\cdots,U\}$. 

During the training stage, a glance language model \cite{DBLP:conf/acl/QianZBWQ0YL20} based sampler module is utilized to strengthen the NAR decoder by modeling token inter-dependency. Specifically, the parallel decoder takes in $\boldsymbol{E}^{\mathrm{acou}}$ and $\boldsymbol{H}$ to generate the first pass decoding result $\mathbf{P(Y)}$. Then, the greedy search is performed to obtain the first pass hypothesis $\mathrm{HYP}'(\mathbf{Y})$. The Hamming distance $d$ between $\mathrm{HYP}'(\mathbf{Y})$ and the ground-truth labels $\mathbf{Y}$ is calculated. Subsequently, Paraformer processes the tokens in $\mathbf{Y}$ through the embedding layer to obtain dense token representations $\boldsymbol{E}^{\mathrm{tok}}=\{\boldsymbol{e}^{\mathrm{tok}}_{u} \mid u=1,\cdots,U\}$. Then, the sampler randomly selects the $\boldsymbol{E}^{\mathrm{tok}}$ with a computed ratio, and the chosen token embeddings are used to replace the acoustic embeddings in $\boldsymbol{E}^{\mathrm{acou}}$, forming the sampled embedding sequence $\boldsymbol{E}^{\mathrm{s}}=\{\boldsymbol{e}^{\mathrm{s}}_u \mid u=1,\cdots,U\}$. Finally, the speech embedding sequence $\boldsymbol{H}$ and $\boldsymbol{E}^{\mathrm{s}}$ are fed into the parallel decoder again to produce the second pass decoding results $\mathbf{P}''(\mathbf{Y})$. The cross-entropy (CE) loss, mean absolute error (MAE), and minimum word error rate (MWER) \cite{DBLP:conf/icassp/PrabhavalkarSWN18} losses are employed to train the Paraformer jointly. The second pass hypotheses are sampled to generate negative candidates for the MWER training, and MAE is computed between the predicted token count and its reference label $U$.
\subsection{Multi-level Semantic Correlation Predictor}
\label{chap:MSG}
Fig.~\ref{fig:overview}(b) describes the architecture of the multi-level semantic correlation predictor (SemCorPredict). During the training of SemCorPredict, the ``Token Embed'' layer encodes the $l$-th token $\mathit{z_{m,l}}$ in the $m$-th biasing phrase $\boldsymbol{z}_m=\{\mathit{z}_{m,l} \mid l=1,\cdots,L\}$, which has a length of $L_m$ and is derived from a given biasing phrase list $\boldsymbol{Z}$. Then, a long short-term memory (LSTM) \cite{DBLP:journals/neco/HochreiterS97} based bias encoder processes all token embeddings of one phrase and outputs the phrase-level embedding $\boldsymbol{e}^{\mathrm{phr}}_m$. Among the biasing list $\boldsymbol{Z}$, a no-bias token ``\textless no-bias\textgreater'' is involved as an option to indicate the absence of context.
Subsequently, a cross-attention module takes in the ``Query'' embeddings $\boldsymbol{E}^{\mathrm{acou}}$, ``Key'' and ``Value'' embeddings $\boldsymbol{E}^{\mathrm{phr}}$ to produce the relevant biased embeddings $\boldsymbol{E}^{\mathrm{bias}}$:
  \begin{align}
   \boldsymbol{E}^{\mathrm{bias}} &= \operatorname{CrossAttn}(\boldsymbol{E}^{\mathrm{acou}},\boldsymbol{E}^{\mathrm{phr}},\boldsymbol{E}^{\mathrm{phr}})
\end{align}
Then, $\boldsymbol{E}^{\mathrm{bias}}$ and $\boldsymbol{E}^{\mathrm{acou}}$ are added to obtain the compound embeddings $\boldsymbol{E}^{\mathrm{comp}}=\{\boldsymbol{e}^{\mathrm{comp}}_u \mid u=1,\cdots,U\}$.

The list-level correlation prediction is regarded as a binary classification task at each decoding step $u$, in which the list-level scorer takes in the acoustic embeddings $\boldsymbol{e}^{\mathrm{acou}}_u$ and $\boldsymbol{e}^{\mathrm{comp}}_u$ to predict the correlation $\mathrm{Q}^{\mathrm{list}}_u$, indicating whether $\boldsymbol{e}^{\mathrm{acou}}_u$ is related to $\boldsymbol{E}^{\mathrm{phr}}$. In practice, biasing phrases may not appear in all sentences, and their length is relatively limited compared to the entire transcription, leading to label imbalance. To alleviate this issue, we apply the focal loss \cite{DBLP:conf/iccv/LinGGHD17} on the prediction and its reference label $\mathbf{Y}^{\mathrm{list}}=\{\mathrm{y}^{\mathrm{list}}_u \in \{0,1\}\mid u=1,\cdots,U\}$:
\begin{align}
  \tau^{\mathrm{list}}_u &= \mathrm{Q}^{\mathrm{list}}_u \mathrm{y}^{\mathrm{list}}_u + (1-\mathrm{Q}^{\mathrm{list}}_u)(1-\mathrm{y}^{\mathrm{list}}_u) \\ 
  \theta_u &= \alpha \mathrm{y}^{\mathrm{list}}_u + (1-\alpha)(1-\mathrm{y}^{\mathrm{list}}_u) \\
  \mathcal{L}_{\mathrm{list}} &= \sum_{u=1}^{U}(-\theta_u (1-\tau^{\mathrm{list}}_u)^{\gamma} \log(\tau^{\mathrm{list}}_u))
\end{align}
where $\alpha$, ranging from 0 to 1, is used to balance positive and negative labels. $\gamma$ is used to modulate the factor ($1 - \tau^{\mathrm{list}}_u$) and balance easy and hard labels.

In the phrase-level correlation prediction, each biasing phrase is treated as an entity for classification. 
To simplify this classification at the training stage, we assume that each sentence contains only one biasing phrase. In the phrase-level scorer, the element-wise product between the list-level reference label $\mathbf{Y}^{\mathrm{list}}$ and $\boldsymbol{E}^{\mathrm{bias}}$ is calculated first. Then, the product is summed over $u$ to obtain a representation $\boldsymbol{e}^{\mathrm{phr}^{\prime}}$:
\begin{equation}
  \boldsymbol{e}^{\mathrm{phr'}}=\sum_{u=1}^{U}{\mathrm{y}_{u}^{\mathrm{list}}\boldsymbol{e}_u^{\mathrm{bias}}}
  \label{eq:phrase-level1}
\end{equation}
The cosine similarities between $\boldsymbol{e}^{\mathrm{phr'}}$ and all biasing phrase embeddings $\boldsymbol{E}^{\mathrm{phr}}$ are finally calculated:
\begin{align}
  \mathit{s}^{\mathrm{phr}^{\prime}}_{m}&=\frac{<\boldsymbol{e}^{\mathrm{phr'}},\boldsymbol{e}^{\mathrm{phr}}_{m}>}{\|\boldsymbol{e}^{\mathrm{phr'}}\|_2\|\boldsymbol{e}^{\mathrm{phr}}_{m}\|_2} \label{eq:phrase-level2} 
\end{align}
To optimize the SemCorPredict, contrastive loss \cite{DBLP:conf/nips/KhoslaTWSTIMLK20} is utilized to maximize the similarity within the target biasing phrase and minimize its similarity with negative biasing phrases:
\begin{align}
   \mathcal{L}_{\mathrm{phr}}&=\sum_{m=1}^{M}\{-\mathit{s}^{\mathrm{phr'}}_{m}~\mathrm{y}^{\mathrm{phr}}_{m}+\mathit{s}^{\mathrm{phr'}}_m(1-\mathrm{y}^{\mathrm{phr}}_m)\}
\end{align}
where $\mathrm{y}^{\mathrm{phr}}_m$ denotes the reference label of the $m$-th phrase, and $\mathbf{Y}^{\mathrm{phr}}=\{\mathrm{y}^{\mathrm{phr}}_m \in \{0,1\} \mid m=1,\cdots,M\}$ denotes the full label set. It is worth noting that the biasing-phrase classification serves as an auxiliary task to guide the attention module training. At the inference stage, due to the absence of the list-level label $\mathbf{Y}^{\mathrm{list}}$, the phrase-level correlation at the $u$-th decoding step is extracted from the multi-head attention matrix $\boldsymbol{A} \in \mathbb{R}^{U \times M \times N}$ ($N$ is the number of heads) of the cross-attention module:
\begin{align}
    \mathbf{Q}^{\mathrm{phr}}_{u} &= \max_{n\in[1,N]}(\boldsymbol{A}_{u,:,n})
\end{align}

The token-level correlation prediction is similar to the decoding process of the ASR backbone. The token-level scorer shares the same architecture as the parallel decoder of the ASR backbone.
Their differences are that the token-level scorer uses $\boldsymbol{E}^{\mathrm{comp}}$ instead of $\boldsymbol{E}^{\mathrm{acou}}$ and has an extra feed-forward output layer. The SemCorPredict can be optimized by the cross-entropy loss $\mathcal{L}_{\mathrm{tok}}$ between the correlation distribution of token classification $\mathbf{Q}^{\mathrm{tok}}$ and its reference label $\mathbf{Y}$.

The ASR backbone is trained first, then the SemCorPredict is jointly trained with three loss functions, i.e., focal loss $\mathcal{L}_{\mathrm{list}}$, contrastive loss $\mathcal{L}_{\mathrm{phr}}$ and cross-entropy loss $\mathcal{L}_{\mathrm{tok}}$:
\begin{align}
  \mathcal{L}_{\mathrm{SemCorPredict}}&=\mathcal{L}_{\mathrm{list}}+\mathcal{L}_{\mathrm{phr}}+\mathcal{L}_{\mathrm{tok}}
\end{align}
During the training of SemCorPredict, the open-source tool Spacy\footnote{\url{https://spacy.io/models/zh#zh_core_web_lg}} is utilized to perform textual parsing on each transcription of the training set. The semantically coherent nouns are randomly selected to form the biasing list, with a probability of 90\% for a sentence to be chosen with a biasing phrase.
Fig.~\ref{fig:overview}(b) shows a simple case that ``Beijing'' is the $m$-th biasing phrase in the given biasing list and it appears in the transcription $\mathbf{Y}$, thus the indices among $u$ of ``Beijing'' in $\mathbf{Y}^{\mathrm{list}}$ are set to 1 and the $m$-th value in $\mathbf{Y}^{\mathrm{phr}}$ is set to 1. 

\subsection{Semantic Correlation Joint Modeling Method}
\label{chap:semchain}
The cross-granularity consistency of the most relevant biasing information is intuitive and intrinsic, making it possible to jointly model the intersection of the three semantic correlations.
Specifically, at the $u$-th decoding step, if an ASR intermediate representation $\boldsymbol{e}^{\mathrm{acou}}_{u}$ shares consistent information with any biasing phrase $\boldsymbol{e}^{\mathrm{phr}}_{m}$, the list-level correlation $\mathrm{Q}^{\mathrm{list}}_u$ becomes prominent, and there must be corresponding phrase and ASR vocabulary tokens that exhibit higher phrase-level correlation $\mathbf{Q}^{\mathrm{phr}}_{u}$ and token-level correlation $\mathbf{Q}^{\mathrm{tok}}_{u}$. Moreover, if certain biasing information shows higher correlations across all three semantic granularities simultaneously, it will achieve the highest correlation within the intersected correlations, making it the most relevant biasing information. An overview of the joint modeling process for multiple semantic correlations is presented in Eq.~(\ref{eq:jointmodel}).

To improve the stability of the predicted $\mathbf{Q}^{\mathrm{list}}$ and $\mathbf{Q}^{\mathrm{phr}}$, two different sliding window smoothing methods, triangular window smoothing, and list-correlation-guided smoothing, are proposed. The smoothed list-level correlation $\mathbf{Q}^{\mathrm{slist}}$ is calculated as below:
\begin{align}
  \mathbf{Q}^{\mathrm{slist}} = \textnormal{Conv1d}(\mathbf{Q}^{\mathrm{list}}, \boldsymbol{\mathcal{W}}^{\mathrm{tri}})
\end{align}
where $\boldsymbol{\mathcal{W}}^{\mathrm{tri}}=[\frac{1-\omega}{2}, \omega, \frac{1-\omega}{2}]$ and the hyperparameter $\omega$ belongs to $[0,1]$. When designing the sliding window for the phrase-level correlation $\mathbf{Q}^{\mathrm{phr}}$, the length of the possible biasing phrase should be considered, thus we propose the list-correlation-guided smoothing method. Specifically, $\mathbf{Q}^{\mathrm{slist}}$ is summed among $u$ to estimate the length $L'$ of the possible biasing phrase, which is then used as the window size. At each output step $u$, the token could appear at any position within the potential biasing phrase. The position of the biasing phrase is determined by finding the sliding window with the highest sum of $\mathbf{Q}^{\mathrm{slist}}$ that includes the token. The calculation process of the list-correlation-guided smoothing is as follows: 
  \begin{align}
    L'&=\textnormal{round}(\sum_{u=1}^{U}\mathbf{Q}^{\mathrm{slist}}_u) \\
    j_{u}&=\argmax_{j\in[u-L'+1, u+L'-1]}(\textnormal{Conv1d}(\mathbf{Q}^{\mathrm{list}}, \mathbbm{1}^{L'})[j])
  \end{align}
where $\mathbbm{1}^{L'}$ is a one-dimensional convolution kernel of length $L'$ with weights of 1, $j_u$ is the index of the most relevant location for the potential biasing phrase. This location determines the corresponding smoothed phrase-level correlation $\mathbf{Q}^{\mathrm{sphr}}_u$ associated with the ASR intermediate representation, the index selection function, $\textnormal{IdxSel}$, indexes the $\mathbf{Q}^{\mathrm{sphr}}_u$ using $j_{u}$:
\begin{align}
  \mathbf{Q}^{\mathrm{sphr}} &= \textnormal{Conv2d}(\mathbf{Q}^{\mathrm{phr}}, \mathbbm{1}^{L'\times 1}) \\
  \mathbf{Q}^{\mathrm{sphr}}_u &= \mathbf{Q}^{\mathrm{sphr}}[u-L'+1:u+L'-1] \\
  \mathbf{Q}^{\mathrm{sphr}}_u &= \textnormal{tanh}(\textnormal{IdxSel}(\mathbf{Q}^{\mathrm{sphr}}_u),j_{u})
\end{align}
The smoothed phrase-level correlation $\mathbf{Q}^{\mathrm{sphr}}$ becomes sharper after list-correlation-guided smoothing, alleviating the issue of flattened attention weight $\mathbf{Q}^{\mathrm{phr}}$ caused by extremely long biasing lists. 
After smoothing, the intersected correlation $\mathbf{Q}^{\mathrm{bias}}_{u}$ can be modeled:
\begin{equation}
  \begin{aligned}
    &\mathbf{Q}^{\mathrm{bias}}_{u}=\textnormal{Softmax}(\max_{m\in[1,M]}\mathrm{Q}^{\mathrm{slist}}_u~\mathbf{Q}^{\mathrm{sphr}}_{u,m}~\boldsymbol{\Phi}_{m,v}~\mathbf{Q}^{\mathrm{tok}}_{u,v})
  \end{aligned}
\end{equation}

\begin{equation}
\begin{aligned}
  \boldsymbol{\Phi}_{m,v}&= \begin{cases}
    1,~if~\mathrm{y}_{v} \in \boldsymbol{z}_{m}\\
    0,~others
  \end{cases}
\end{aligned}
\label{contextualASR}
\end{equation}
where $\boldsymbol{\Phi}_{m,v}$ is an indicator function representing whether the $v$-th ASR vocabulary token $\mathrm{y}_{v}$ appears in a biasing phrase $\boldsymbol{z}_{m}$.

Whether to introduce contexts can be automatically controlled by list-level correlation interpolating, which is used in the collaborative decoding of the joint modeling result $\mathbf{Q}^{\mathrm{bias}}$ and the backbone recognition result $\mathbf{P}^{\mathrm{bb}}$ to obtain the contextualized result $\mathbf{Q}^{\mathrm{casr}}$:
\begin{equation}
  \begin{aligned}
    \mathbf{Q}^{\mathrm{casr}}_u = (1-\mathrm{Q}^{\mathrm{slist}}_{u}) \mathbf{P}^{\mathrm{bb}}_{u}+\mathrm{Q}^{\mathrm{slist}}_{u} \mathbf{Q}^{\mathrm{bias}}_{u}
    \label{eq:coldec2}
  \end{aligned}
\end{equation}

In real applications, many sentences do not contain any specified biasing phrases in that the introduced biasing information may be harmful to the recognition results, which is termed as over-biasing problem. To alleviate the over-biasing issue, a simple and efficient post-processing strategy is utilized: the contextual hypothesis $\mathrm{HYP}^{\mathrm{casr}}$ is selected as the final output only when it detects more biasing phrases than the backbone hypothesis $\mathrm{HYP}^{\mathrm{bb}}$.
\begin{align}
  \small
  \mathrm{HYP}^{\mathrm{casr}}= \begin{cases}
  \mathrm{HYP}^{\mathrm{casr}},
  \text{if} ~\mathcal{C}(\mathrm{HYP}^{\mathrm{casr}})>\mathcal{C}(\mathrm{HYP}^{\mathrm{bb}}) \\
  \\
    \mathrm{HYP}^{\mathrm{bb}},~\text{others}
  \end{cases}
\end{align}
where $\mathcal{C}$ denotes the counter function which counts the number of the detected biasing phrases in a hypothesis. 

\begin{algorithm}[t]
  \footnotesize
  \caption{Group Competitive Purification}\label{algorithm}
  \setlength{\lineskip}{0.5em}
  \KwData{biasing phrase embeddings $\boldsymbol{E}^{\mathrm{phr}}$, acoustic embeddings $\boldsymbol{E}^{\mathrm{acou}}$, speech representations $\boldsymbol{H}$, purification group size $group\_size$, purification rounds $n\_r$, list-level correlation threshold $thres\_list$, top number $n\_top$}
  \KwResult{purified biasing phrase embeddings $\boldsymbol{E}^{\mathrm{phr,pur}}$}
  \# Using decoding of a sentence as an example.\\
  $i \leftarrow 1$\;
  $G \leftarrow  \lceil M /\ group\_size \rceil$ \\
  \While{$i \leq n\_r~\textnormal{and} ~ G > 1$}{
  $\{\boldsymbol{E}^{\mathrm{phr}}_g\}_{g=1}^G \leftarrow \boldsymbol{E}^{\mathrm{phr}}.\textnormal{reshape}(G, group\_size, :)$  \\
  $\{~~\mathbf{Q}^{\mathrm{list}}_g, \mathbf{Q}^{\mathrm{phr}}_g = \textnormal{SemCorPredict}(\boldsymbol{E}^{\mathrm{acou}},\boldsymbol{E}^{\mathrm{phr}}_g,\boldsymbol{H})~~\}_{g=1}^G$\\
  $\{~~\mathbf{Q}^{\mathrm{list}}_g \in \{0,1\} \leftarrow \text{bool}(\mathbf{Q}^{\mathrm{list}}_g  > thres\_list)~~\}_{g=1}^G$ \\
   $\boldsymbol{I} \leftarrow \text{shuffle}(\text{unique}(\{(\mathbf{Q}^{\mathrm{list}}_g \odot \mathbf{Q}^{\mathrm{phr}}_g).\textnormal{topk}(n\_top)\}_{g=1}^G))$ \\
  $M\leftarrow \textnormal{len}(\boldsymbol{I})$ \\ 
  $\boldsymbol{E}^{\mathrm{phr}} \leftarrow \textnormal{IdxSel}(\boldsymbol{E}^{\mathrm{phr}}, \boldsymbol{I})$ \\
  $G \leftarrow  \lceil M /\ group\_size \rceil$ \\
  $i \leftarrow i+1$\\     
  }
  $M^{\mathrm{pur}} \leftarrow M$ \\
  \Return{$\boldsymbol{E}^{\mathrm{phr}},M^{\mathrm{pur}}$}
  \end{algorithm}

\subsection{Group and Competitive Purification} 
\label{chap:EM}
The purification mechanism is proposed to reduce the computational cost introduced by the joint modeling across three semantic granularities.
As shown in Fig.~\ref{fig:overview}(c), the key idea of purification is that only a few biasing phrases with higher list- and phrase-level correlations are selected in each group for constructing the purified short biasing list. The strategy of dividing biasing phrases into groups based on the principle of divide and conquer, combined with performing parallel purification in group competitive purification (GCP), can yield more accurate semantic correlation prediction, especially when $M$ varies significantly compared to the training condition. Algorithm \ref{algorithm} provides the pseudo-code of the calculation procedure. In each round, the embeddings of biasing list $\boldsymbol{E}^{\mathrm{phr}}$ are grouped randomly into several short lists $\boldsymbol{E}^{\mathrm{phr}}_g$. The SemCorPredict uses $\boldsymbol{E}^{\mathrm{acou}}$, $\boldsymbol{E}^{\mathrm{phr}}_g$ and $H$ to generate grouped list-level correlation $\mathbf{Q}^{\mathrm{list}}_g$ and grouped phrase-level correlation $\mathbf{Q}^{\mathrm{phr}}_g$ in a parallel way. At each decoding step, $\mathbf{Q}^{\mathrm{list}}_{u,g}$ is reset to $\mathbf{1}$ (True) or $\mathbf{0}$ (False) according to the relationship with list-level correlation threshold $thres\_list$. The top $n\_top$ active biasing phrases are the winners at each ``True'' decoding step, and then they are shuffled and re-grouped for the next round.
Notably, the purification running \textbf{once} means that the $group\_size$ equals the length of the biasing list, which is termed once competitive purification (OCP).

\begin{table}
  \caption{Statistics of AISHELL-NER}
  \label{table1}
  \footnotesize
  \centering
  \setlength{\tabcolsep}{0.3mm}
  \renewcommand{\arraystretch}{1.3}
  \begin{tabular}{l|c|c|c|c|c|c|c}
  \hline
  \multirow{2}{*}{Sets}&\multirow{2}{*}{Categ.}&\multirow{2}{*}{\#Sent.}&\multirow{2}{*}{\# Phraselist}&\multirow{1}{*}{\scriptsize Length of phrase}&\multirow{1}{*}{Inclusion}&\multicolumn{2}{c}{OOV Rate (\%)}\\
  \cline{7-8}
  &&&&\scriptsize{min,max,avg}&Rate (\%)&\scriptsize{AISHELL-1}&\scriptsize{KeSpeech}\\
  \hline
  \multirow{4}*{Dev}&NE&\multirow{4}*{14326}&2204&2,19,4&34.94&51.57&49.52\\ 
  \cline{2-2}\cline{4-8}
  &PER&&706&2,9,3&11.20&56.31&65.25\\
  \cline{2-2}\cline{4-8}
  &LOC&&612&2,12,3&15.41&43.86&32.57\\ 
  \cline{2-2}\cline{4-8}
  &ORG&&888&2,19,5&19.11&53.10&48.70\\
  \hline
  \multirow{4}*{Test}&NE&\multirow{4}*{7176}&1196&2,16,4&31.73&50.71&49.12\\
  \cline{2-2}\cline{4-8}
  &PER&&414&2,11,3&\textbf{10.30}&\textbf{62.22}&\textbf{65.38}\\
  \cline{2-2}\cline{4-8}
  &LOC&&343&2,11,3&15.25&38.60&30.99\\
  \cline{2-2}\cline{4-8}
  &ORG&&441&2,16,5&16.97&49.32&47.95\\
  \hline
\end{tabular}
\end{table}
\section{EXPERIMENTS}
\subsection{Data and Metrics}
\noindent In practice, the biasing list may be set arbitrarily for contextual ASR, thus the length of the biasing list and the number of tokens per phrase can vary significantly.
To simulate this situation, we employ an open-source Chinese named entity data set called AISHELL-NER \cite{DBLP:conf/icassp/ChenXWXZH22} to form the biasing list for evaluation, in which a named entity is viewed as a biasing phrase.
A named entity phrase usually contains several words and the token number of a phrase is larger than that of a word. Thus, named entity phrases may be more suitable for the method evaluation than word-level biasing phrases \cite{DBLP:conf/interspeech/LeJKKSMCSFKSS21}.

Table \ref{table1} provides statistics of the biasing information on the development and test sets\footnote{All biasing lists are present at \url{https://github.com/shibeiing/Purified-Semantic-Correlation-Joint-Modeling-PSC-Joint}.}, including the number of sentences, the length of biasing lists, the length statistics of phrases, the percentages of sentences including the biasing phrase (Inclusion Rate), and the rate of out-of-vocabulary (OOV). 
The OOV rate measures the fraction of phrases appearing in the biasing lists but not in the training data. 

The development and test sets of AISHELL-NER both consist of three categories of the named entity: person names (PER), location names (LOC), and organization names (ORG). We combine the three mentioned named entity categories to create the fourth biasing list called ``NE''. As indicated in the fourth and fifth columns of Table~\ref{table1}, the ``NE'' lists contain thousands of biasing phrases, with the number of tokens per phrase (phrase length) ranging from 2 to 19.
To evaluate the impact of varying biasing list lengths, we split the ``NE'' biasing list randomly into several lists with different lengths, e.g., ``NE-101''. 
The lower inclusion rate means that biasing phrases are rarer in the text, i.e., the biasing information is redundant for most sentences.
Thus, the ``PER-414'' biasing list of the test set, which has a higher OOV rate and lower inclusion rate, is also employed to evaluate the performance degradation caused by the over-biasing issue.

To investigate the impact of training data scale for the PSC-Joint approach, experiments are mainly conducted on two open-source Chinese corpora, AISHELL-1 \cite{DBLP:conf/ococosda/BuDNWZ17} and Kespeech \cite{DBLP:conf/nips/Tang0XSLZWTXZYL21}. The AISHELL-1 contains 178 hours of standard Mandarin audio, and the KeSpeech Phase-1 corpus contains 895 hours of transcribed audio in Mandarin and its eight subdialects. The former one is used to perform rapid experimental validation and analysis, and the latter one, which contains many more accented speech utterances, is suitable to verify the generalization of the proposed methods in reality.

In all experiments, we evaluate the performance using the following metrics: character error rate (CER), recall, precision and F1 score. The CER and F1 score, which reflect the trade-off between under- and over-biasing in contextual ASR, are mainly discussed in the results subsection. Notably, recall, precision, and F1 score are computed based on exact match, where the entire biasing phrase must be recognized without any token errors.
In addition, the real-time factor (RTF) is used to evaluate the impact of PSC-Joint on the inference speed.

\subsection{Experimental Setup}
Following the official configuration and recipe from FunASR \cite{gao23g_interspeech}, we use 80-dimensional FBANK features as input and Chinese characters as ASR vocabulary tokens to train the ASR backbone and contextual ASR models on the AISHELL-1 and KeSpeech Phase-1 corpora.
The model includes a 12-layer encoder based on Conformer \cite{DBLP:conf/interspeech/GulatiQCPZYHWZW20} and a 6-layer parallel decoder based on Transformer \cite{DBLP:conf/nips/VaswaniSPUJGKP17}.
The parallel decoder consists of multiple blocks stacked with self-attention, feedforward networks \cite{DBLP:conf/interspeech/GaoZLM20}, and cross-attention. The cross-attention module incorporates a transformer-style multi-head attention layer with an embedding size of 256 and four attention heads. More details can be found in the recipe\footnote{\url{https://github.com/alibaba-damo-academy/FunASR/tree/v0.2.0/egs/aishell/paraformer/conf}}. The token-level scorer closely resembles the parallel decoder in the baseline, with the sole distinction of an additional feed-forward layer inserted before the output layer.
The bias encoder consists of a single-layer LSTM with 256 hidden units. The token embedding layer shares parameters with the pre-trained embedding layer from the ASR backbone. 
In the list-level scorer, the dimension of the feed-forward layer is set to 512. The phrase-level scorer does not contain trainable parameters, and its calculation process is defined in Eq. (\ref{eq:phrase-level1}) and Eq. (\ref{eq:phrase-level2}).

Considering the extreme imbalance between positive and negative labels in the list-level correlation, the hyperparameters $\alpha$ and $ \gamma$ of focal loss are set to 0.75 and 2.0, respectively. During inference, the hyperparameter $\omega$ of triangular window smoothing is set to 0.6. In the purification mechanism, the size of group $group\_size$, the number of rounds $n\_r$, the threshold for list-level correlation $thres\_list$, and the number of biasing phrases being kept at each group $n\_top$ are set to 75, 2, 0.5, 10, respectively. Note that the parameters of the pre-trained ASR backbone are frozen during the training stage of SemCorPredict. SpecAugment \cite{DBLP:conf/interspeech/ParkCZCZCL19} is applied for the data augmentation of the AISHELL-1 task. Following the configuration of FunASR \cite{gao23g_interspeech}, the ASR backbone of the AISHELL-1 task is trained for 50 epochs. Considering the scale of the data, the training epoch of the KeSpeech ASR backbone model is set to 150. According to preliminary experiments, the SemCorPredict module is trained for 50 epochs on both AISHELL-1 and KeSpeech tasks.


\begin{table*}[t!]
  \scriptsize
  \caption{\footnotesize Comparison of Contextual ASR Models Tested on Varying-length Lists: \textbf{CER // Recall\textbar{}Precision\textbar{}F1 score} (\%). (AISHELL-1)}
  \label{tab:methodcomparitionAishell}
  \centering
  \setlength{\tabcolsep}{0.8mm}
  \renewcommand{\arraystretch}{0.6}
  \begin{tabular}{l|c|c|c|c|c|c|c|c|c}
    \toprule
    \textnormal{Models}
    &\textnormal{Pur.}
     &\multicolumn{1}{c|}{\textnormal{NE-51}}
     &\multicolumn{1}{c|}{\textnormal{NE-101}} 
     &\multicolumn{1}{c|}{\textnormal{NE-201}}
     &\multicolumn{1}{c|}{\textnormal{NE-401}}
     &\multicolumn{1}{c|}{\textnormal{NE-601}}
     &\multicolumn{1}{c|}{\textnormal{NE-801}}
     &\multicolumn{1}{c|}{\textnormal{NE-1001}}
     &\multicolumn{1}{c}{\textnormal{NE-1196}} \\
    \midrule
    \multirow{2}{*}{Baseline} & \multirow{2}{*}{-} &5.05&5.05&5.05 &5.05 & 5.05 & 5.05 & 5.05 & 5.05 \\
     &  &76.36\textbar{}\textbf{100.0}\textbar{}86.60& 71.49\textbar{}\textbf{99.41}\textbar{}83.17&77.13\textbar{}\textbf{99.07}\textbar{}86.73  &77.99\textbar{}\textbf{99.68}\textbar{}87.51 & 83.08\textbar{}\textbf{99.40}\textbar{}90.51 &80.21\textbar{}\textbf{99.54}\textbar{}88.83& 77.75\textbar{}\textbf{99.31}\textbar{}87.21 &79.90\textbar{}\textbf{99.19}\textbar{}88.50 \\
     \midrule
     \multirow{2}{*}{ColDec \cite{DBLP:conf/icassp/HanDZX21}} & \multirow{2}{*}{\ding{55}} &5.06 & 5.06 & 5.09 & 5.13& 5.14 & 5.16 & 5.17 & 5.16 \\
      &  &84.55\textbar{}98.94\textbar{}\textbf{91.18} & 74.47\textbar{}98.31\textbar{}84.75 & 80.22\textbar{}98.88\textbar{}88.58 & 80.46\textbar{}99.39\textbar{}88.93 & 85.05\textbar{}99.32\textbar{}91.63 & 82.37\textbar{}99.59\textbar{}90.17 & 79.98\textbar{}99.09\textbar{}88.52 & 81.47\textbar{}99.03\textbar{}89.40 \\
    \midrule
    \multirow{2}{*}{FineCoS \cite{DBLP:conf/icassp/HanDLCZMX22}} & \multirow{2}{*}{PS} &5.07 & 5.07 & 5.08 & 5.07& 5.10 & 5.09 & 5.10 & 5.09 \\
     &  &84.55\textbar{}97.89\textbar{}90.73 & 76.60\textbar{}98.36\textbar{}86.12 & 79.49\textbar{}98.65\textbar{}88.04 & 79.80\textbar{}99.08\textbar{}88.40 & 85.05\textbar{}99.08\textbar{}91.53 & 82.11\textbar{}99.23\textbar{}89.87 & 80.01\textbar{}99.01\textbar{}88.50 & 81.52\textbar{}99.03\textbar{}89.43 \\
    \midrule
    \multirow{2}{*}{SeACo \cite{10446106}} & \multirow{2}{*}{ASF} & 5.05& 5.06 & 5.03 & 5.02 & 5.01 & 5.00 & 4.99 & 4.98 \\
    &  & 87.27\textbar{}95.05\textbar{}91.00 & 78.72\textbar{}94.87\textbar{}86.05 & 81.49\textbar{}97.82\textbar{}88.91 & 80.21\textbar{}98.98\textbar{}88.62 & 85.05\textbar{}99.17\textbar{}91.57 & 81.96\textbar{}99.46\textbar{}89.87 & 79.69\textbar{}99.21\textbar{}88.39& 81.40\textbar{}99.03\textbar{}89.35 \\
    \midrule
    \multirow{2}{*}{SC-Joint} & \multirow{2}{*}{\ding{55}} & 5.31 & 5.18 & 5.13 & 5.04~ & 4.95 & 4.95 & 4.87 & 4.80 \\
     &  & \textbf{89.09}\textbar{}90.74\textbar{}89.91 & 85.53\textbar{}93.06\textbar{}\textbf{89.14} & 85.12\textbar{}98.32\textbar{}91.25 & 85.57\textbar{}98.48\textbar{}91.57 & 88.05\textbar{}99.01\textbar{}93.21 & 85.10\textbar{}99.13\textbar{}91.58 & 83.65\textbar{}99.05\textbar{}90.70&84.98\textbar{}98.90\textbar{}91.41 \\
     \midrule
    \multirow{2}{*}{SC-Joint-P} & \multirow{2}{*}{\ding{55}} & \textbf{5.04} & \textbf{5.02} & \textbf{5.00} & 4.96 & 4.93 & 4.92 & 4.86& 4.84 \\
     & & \textbf{89.09}\textbar{}90.74\textbar{}89.91 & 85.53\textbar{}93.06\textbar{}\textbf{89.14} & 85.12\textbar{}98.32\textbar{}91.25 & 85.66\textbar{}98.58\textbar{}91.66 & 88.22\textbar{}99.02\textbar{}93.31 & 85.32\textbar{}99.13\textbar{}91.71 & 83.84\textbar{}99.06\textbar{}90.82 & 85.20\textbar{}98.90\textbar{}91.54 \\
    \midrule
    \multirow{2}{*}{PSC-Joint} & \multirow{2}{*}{OCP} & 5.41 &  5.32 &  5.32 &  5.23& 5.14 &  5.10~ &  4.95 & 4.89 \\
     &  & \textbf{89.09}\textbar{}83.76\textbar{}86.34  &  \textbf{86.38}\textbar{}91.86\textbar{}89.04 &  86.39\textbar{}96.95\textbar{}91.36& 86.48\textbar{}98.13\textbar{}91.94 & 89.10\textbar{}98.52\textbar{}93.57 &  86.26\textbar{}98.38\textbar{}91.92 & 85.41\textbar{}98.60\textbar{}91.53 & 86.24\textbar{}98.54\textbar{}91.98 \\
     \midrule
    \multirow{2}{*}{PSC-Joint-P} & \multirow{2}{*}{OCP} &  5.05 &  5.03&  5.07 & 4.95  &  \textbf{4.92} &  4.90 &  4.82 &  4.79\\
    &  &  \textbf{89.09}\textbar{}83.76\textbar{}86.34&  \textbf{86.38}\textbar{}91.86\textbar{}89.04&  86.57\textbar{}96.95\textbar{}91.47 & 86.64\textbar{}98.22\textbar{}92.07  &  89.31\textbar{}98.53\textbar{}93.69 &  86.41\textbar{}98.38\textbar{}92.01 & 85.70\textbar{}98.60\textbar{}91.70 & 86.48\textbar{}98.59\textbar{}92.14\\
    \midrule
    \multirow{2}{*}{PSC-Joint} & \multirow{2}{*}{GCP} &5.31 &  5.26 &  5.28 & 5.20  &  5.15 &  5.10 &  4.98 & 4.87 \\
     &  &\textbf{89.09}\textbar{}90.74\textbar{}89.91  &  \textbf{86.38}\textbar{}91.03\textbar{}88.65 &  86.93\textbar{}96.57\textbar{}91.50& 87.30\textbar{}97.60\textbar{}92.17  &  89.89\textbar{}98.40\textbar{}93.95 &  87.19\textbar{}98.15\textbar{}92.35 &  85.82\textbar{}98.28\textbar{}91.63 & 86.99\textbar{}98.52\textbar{}92.39 \\
    \midrule
     \multirow{2}{*}{PSC-Joint-P}&\multirow{2}{*}{GCP} &\textbf{5.04}  &  5.03 & \textbf{5.00} & \textbf{4.94}  & \textbf{4.92}  &  \textbf{4.88} &  \textbf{4.81}& \textbf{4.76} \\
      & &\textbf{89.09}\textbar{}90.74\textbar{}89.91 &  \textbf{86.38}\textbar{}91.03\textbar{}88.65 & \textbf{87.11}\textbar{}96.58\textbar{}\textbf{91.60}  & \textbf{87.39}\textbar{}97.70\textbar{}\textbf{92.25}  & \textbf{90.02}\textbar{}98.40\textbar{}\textbf{94.02} &  \textbf{87.34}\textbar{}98.15\textbar{}\textbf{92.43} &  \textbf{86.08}\textbar{}98.29\textbar{}\textbf{91.78}& \textbf{87.18}\textbar{}98.52\textbar{}\textbf{92.51} \\
    \bottomrule
  \end{tabular}
\end{table*}

\subsection{Results on AISHELL-1}
\noindent Table \ref{tab:methodcomparitionAishell} compares the baseline, three non-autoregressive contextual ASR models and the PSC-Joint model, all trained on the AISHELL-1 corpus. The baseline does not utilize the biasing information and serves as the backbone of other contextual models. For a fair comparison, ColDec, FineCoS, SeACo, and the PSC-Joint models share the same pre-trained backbone, and all these models conduct collaborative decoding with it through interpolation weighting. 
The ColDec, FineCoS and SeACo models are trained using a single contextual cross-entropy loss, and the biasing phrase is incorporated through a single bias probability.
This can be regarded as using a single probability to represent list-level, phrase-level, and token-level correlations, serving as a baseline to highlight the effectiveness of our explicit supervision of three semantic correlations and their joint modeling.
To purify the biasing list, the FineCoS model uses phrase selection (PS), while the SeACo model employs attention score filtering (ASF). The ``SC-Joint'' model indicates that no purification mechanism is applied, and models with the ``-P'' suffix incorporate the post-processing discussed in the Section \ref{chap:semchain}. To assess the robustness of models to varying-length lists, results on several biasing lists are reported. 

\begin{figure}[t!]
	\centering
	\includegraphics[width=\linewidth]{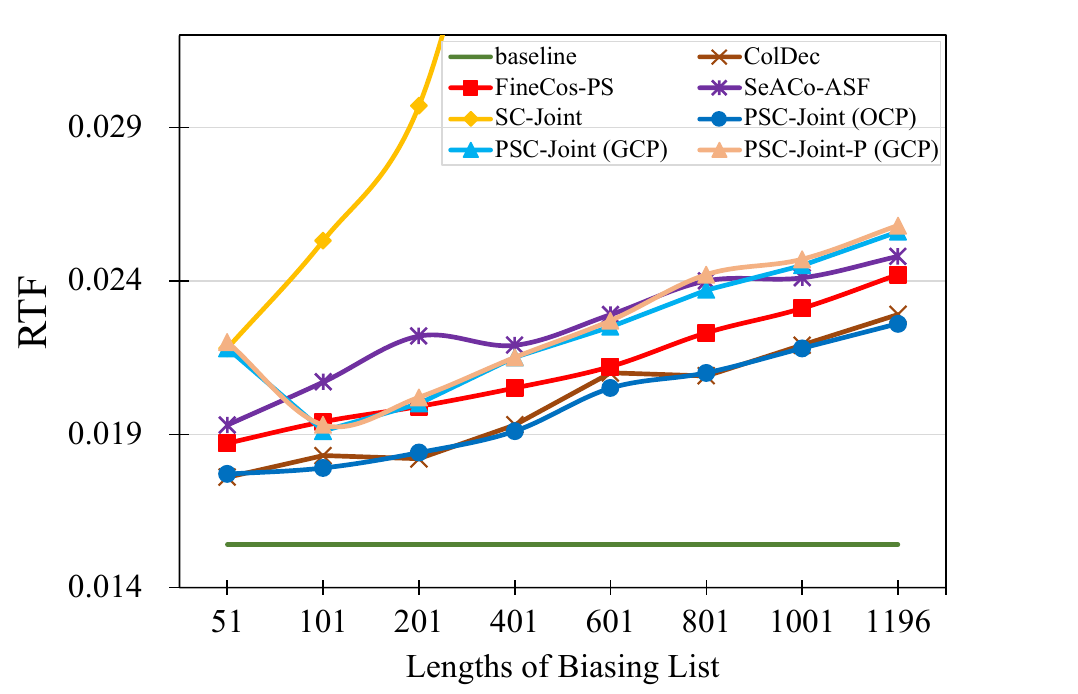}
	\caption{Comparison of contextual ASR models tested on varying-length biasing lists: RTF.}
	\label{fig:rtf}
\end{figure} 

Comparing each column in Table \ref{tab:methodcomparitionAishell}, it can be observed that almost all contextual ASR models surpass the backbone model on F1 score, and the PSC-Joint-P (GCP) model achieves 25.57\%, 18.23\%, and 18.64\% relative F1 score improvement over ColDec, FineCoS, and SeACo on the ``NE-101'' list, respectively. As the biasing list grows longer, it becomes more challenging to detect biasing phrases. Compared with the ColDec model, the SC-Joint model performs better on F1 score as the length of the biasing lists increases, demonstrating that the proposed semantic correlation joint modeling method can significantly enhance the robustness of contextual ASR against changes in the biasing information volume.
Even with PS and ASF purification methods, the performance improvements of the FineCoS and SeACo models decrease when the length of the list increases. Nevertheless, PSC-Joint-P (GCP) achieves average F1 score improvements of 23.08\% and 21.99\% across varying-length biasing lists compared with FineCoS and SeACo. Assisted by the post-processing strategy, PSC-Joint-P (GCP) achieves lower CER on varying-length biasing lists while maintaining the F1 score improvement.



Fig.~\ref{fig:rtf} illustrates the real-time factor (RTF) of the baseline and contextual ASR models tested on varying-length biasing lists. When comparing the inference speed of the PSC-Joint (GCP) model and the PSC-Joint-P (GCP) model, we observe that the negative effect caused by the post-processing is negligible, indicating that the post-processing alleviates the over-biasing problem at a low cost. In addition, the inference speeds of the FineCoS, SeACo, and PSC-Joint-P (GCP) are almost the same, and their RTFs increase nearly linearly with the list length.

We further conduct a modular analysis on PSC-Joint-P with the results in Table~\ref{tab:methodcomparitionAishell} and Fig.~\ref{fig:rtf}. Despite lacking purification, SC-Joint exhibits robustness across varying list lengths, which benefits from the joint modeling across multi-level semantic correlations.
However, as the length of the biasing list increases, the RTF of the SC-Joint increases significantly.
Compared with SC-Joint, PSC-Joint (OCP) and PSC-Joint (GCP) benefit from the purification, achieving superior performance and faster inference speed. 

A potential risk of applying purification to biasing lists is the unintended removal of a target biasing phrase that appears in the text sentence. 
To compare the OCP and GCP mechanisms, we introduce a new metric, retention rate, which indicates the average proportion of target biasing phrases in each test sentence that are successfully retained in the purified phrase list after the purification process.
The random grouping strategy, based on the principle of divide and conquer, reduces the competition among similar phrases at the beginning of purification. Although it introduces some additional computations, GCP achieves a retention rate of 92.06\% on the ``NE-1196'' biasing list, outperforming the 86.26\% retention rate achieved by OCP. Therefore, PSC-Joint (GCP) performs better across varying-length biasing lists than PSC-Joint (OCP), benefiting from the divide-and-conquer strategy.

\subsection{Ablation Study}
Table~\ref{tab:ablation} presents an ablation study of the PSC-Joint model, evaluating the effectiveness of multi-level semantic correlation losses on biasing lists of different lengths. To perform dynamic interpolation (defined in Eq.~(\ref{eq:coldec2})), the list-level loss must be retained. Considering that the token- and phrase-level correlation can be derived from the backbone recognition results and the attention module, respectively, their corresponding losses may optionally be excluded during training. Specifically, E1-E4 and E5-E8 present the ablation results of SC-Joint without or with purification, respectively.

We begin by analyzing the effect of token- and phrase-level losses without GCP. Compared to E1, removing the phrase-level loss (E2) leads to a noticeable precision drop (from 98.32\% to 90.75\% on NE-201) and a CER increase (from 4.95\% to 5.18\% on NE-601), suggesting the emergence of over-biasing errors. In contrast, removing the token-level loss while retaining the phrase-level loss (E3) improves precision (from 98.90\% to 99.02\% on NE-1196) but reduces recall (from 85.12\% to 84.75\% on NE-201), indicative of under-biasing.
When both token- and phrase-level losses are removed (E4), model performance becomes unstable across biasing lists and metrics. For instance, while CER consistently increases compared to E1 (4.95\% to 5.05\% on NE-601), F1 scores show mixed results, decreasing on NE-201 (91.25\% to 90.43\%) but slightly increasing on NE-601 and NE-1196. These results reflect the difficulty of balancing over-biasing and under-biasing without the guidance of token- and phrase-level supervision.

\begin{table}[t!]
  \caption{Ablation Study of PSC-Joint Model Tested on Varying-length Biasing Lists: \textbf{CER // Recall\textbar{}Precision\textbar{}F1 score} (\%).}
  \label{tab:ablation}
  \scriptsize
  \centering
  \setlength{\tabcolsep}{0.7mm}
  \renewcommand{\arraystretch}{1.0}
  \begin{tabular}{l|c|c|c|c|c|c}
    \toprule
    \multirow{2}{*}{Exp.}
    &\multirow{1}{*}{\textnormal{Token}}
    &\multirow{1}{*}{\textnormal{Phrase}}
    &\multirow{2}{*}{\textnormal{GCP}} 
     &\multirow{2}{*}{\textnormal{NE-201}}
     &\multirow{2}{*}{\textnormal{NE-601}}
     &\multirow{2}{*}{\textnormal{NE-1196}} \\
      & level & level& & & & \\
    \midrule
     \multirow{2}{*}{E1}& \multirow{2}{*}{\ding{51}}& \multirow{2}{*}{\ding{51}}&& 5.13 &  4.95 & 4.80 \\
       & & &&85.12\textbar{}98.32\textbar{}91.25 & 88.05\textbar{}99.01\textbar{}93.21 &84.98\textbar{}98.90\textbar{}91.41 \\
       \midrule
       \multirow{2}{*}{E2} & \multirow{2}{*}{\ding{51}}&& & 5.34&  5.18 & 4.91 \\
       & & & &87.30\textbar{}90.75\textbar{}88.99 & 89.93\textbar{}97.95\textbar{}93.77 &86.57\textbar{}98.51\textbar{}92.16 \\
        \midrule
        \multirow{2}{*}{E3} & & \multirow{2}{*}{\ding{51}}&& 5.09&  4.97 & 4.90 \\
        & & && 84.75\textbar{}98.11\textbar{}90.94 & 86.22\textbar{}99.23\textbar{}92.27 &83.04\textbar{}99.02\textbar{}90.33 \\
         \midrule
         \multirow{2}{*}{E4}& & & &5.41 &  5.05 & 4.81\\
          & & & &87.48\textbar{}93.59\textbar{}90.43 & 89.52\textbar{}98.35\textbar{}93.72 &85.70\textbar{}98.77\textbar{}91.78 \\
          \midrule
           \midrule
           \multirow{2}{*}{E5}& \multirow{2}{*}{\ding{51}}& \multirow{2}{*}{\ding{51}}&\multirow{2}{*}{\ding{51}}& 5.28&  5.15 & 4.87 \\
    & & &&86.93\textbar{}96.57\textbar{}91.50 & 89.89\textbar{}98.40\textbar{}93.95 &86.99\textbar{}98.52\textbar{}92.39 \\
    \midrule
    \multirow{2}{*}{E6} & \multirow{2}{*}{\ding{51}}&& \multirow{2}{*}{\ding{51}}& 5.42&  5.50 & 5.30 \\
      & & & &87.11\textbar{}90.74\textbar{}88.89 & 90.23\textbar{}97.25\textbar{}93.61 &86.77\textbar{}97.61\textbar{}91.87 \\
      \midrule
      \multirow{2}{*}{E7} & & \multirow{2}{*}{\ding{51}}&\multirow{2}{*}{\ding{51}}& 5.20&  5.04 & 4.85 \\
      & & && 85.48\textbar{}96.12\textbar{}90.49 & 88.97\textbar{}98.34\textbar{}93.42 &86.36\textbar{}98.59\textbar{}92.07 \\
      \midrule
      \multirow{2}{*}{E8} & & & \multirow{2}{*}{\ding{51}}&5.60 &  5.38 & 5.14\\
      & & & &88.75\textbar{}91.23\textbar{}89.97 & 90.31\textbar{}97.12\textbar{}93.59 &87.28\textbar{}98.04\textbar{}92.35\\
    \bottomrule
  \end{tabular}
\end{table}

The challenge is further amplified when introducing the GCP, which relies heavily on predicted list- and phrase-level correlations. Under this setting, the removal of both token- and phrase-level losses (E8) leads to considerable performance degradation compared to E5 (F1 drops from 91.50\% to 89.97\% and CER rises from 5.28\% to 5.60\% on NE-201), highlighting the negative effect of inaccurate biasing phrase selection on purification performance.
In summary, the results clearly demonstrate that (1) token- and phrase-level losses are complementary for mitigating over- and under-biasing, and (2) their presence is essential for fully exploiting the purification capability of GCP. 

In addition, GCP retains a small number of biasing phrases that are highly correlated with ASR intermediate representations, leading to over-biasing, which slightly degrades CER and precision. This issue can be mitigated by the proposed post-processing strategy, as shown in the last row of Table~\ref{tab:methodcomparitionAishell}.

\begin{table*}[t!]
  \caption{Comparison of Contextual ASR Models Tested on the varying-length biasing lists of the development and test sets: \textbf{CER // Recall\textbar{}Precision\textbar{}F1} (\%). (KeSpeech task)}
  \label{tab:methodcomparitionkespeech}
  \scriptsize
  \centering
  \setlength{\tabcolsep}{0.7mm}
  \renewcommand{\arraystretch}{0.6}
  \begin{tabular}{l|c|c|c|c|c|c|c|c|c}
    \toprule
    \multirow{2}{*}{\textnormal{Models}} 
    &\multirow{2}{*}{\textnormal{Pur}}
    & \multicolumn{4}{c|}{\textnormal{AISHELL-NER Dev}} & \multicolumn{4}{c}{\textnormal{AISHELL-NER Test}} \\
    \cmidrule{3-7}\cmidrule{7-10}
    &&\multicolumn{1}{c|}{\textnormal{NE-51}}
    &\multicolumn{1}{c|}{\textnormal{NE-801}} 
    &\multicolumn{1}{c|}{\textnormal{NE-1601}}
    &\multicolumn{1}{c|}{\textnormal{NE-2204}}
    &\multicolumn{1}{c|}{\textnormal{NE-51}}
    &\multicolumn{1}{c|}{\textnormal{NE-601}}
    &\multicolumn{1}{c|}{\textnormal{NE-1196}} 
    &\multicolumn{1}{c}{\textnormal{PER-414}} \\
    \midrule
    \multirow{2}{*}{\textnormal{Baseline}}&\multirow{2}{*}{-}&4.76 & 4.76&4.76& 4.76 & 5.07 & 5.07 & 5.07 & 5.07\\
    &&79.83\textbar{}\textbf{100.0}\textbar{}88.78 & 80.66\textbar{}\textbf{98.83}\textbar{}88.83&78.17\textbar{}\textbf{98.60}\textbar{}87.20& 79.40\textbar{}\textbf{98.81}\textbar{}88.05 & 71.82\textbar{}\textbf{100.0}\textbar{}83.60 & 81.33\textbar{}\textbf{99.44}\textbar{}89.48 & 77.89\textbar{}\textbf{99.05}\textbar{}87.20 & 45.51\textbar{}\textbf{96.43}\textbar{}61.83 \\
    \midrule
    \multirow{2}{*}{ColDec \cite{DBLP:conf/icassp/HanDZX21}} & \multirow{2}{*}{\ding{55}}& 4.77 & 4.75 & 4.74 & 4.74 & 5.07 & 5.01& 4.98 & 4.96 \\
     && 90.56\textbar{}100.0\textbar{}95.05 & 85.47\textbar{}98.33\textbar{}91.45 & 82.22\textbar{}98.28\textbar{}89.54 & 82.41\textbar{}98.61\textbar{}89.78 & 86.36\textbar{}97.94\textbar{}91.79 & 86.59\textbar{}98.95\textbar{}92.36 & 82.17\textbar{}98.78\textbar{}89.71 & 61.69\textbar{}97.34\textbar{}75.52 \\
    \midrule
    \multirow{2}{*}{\textnormal{FineCoS~\cite{DBLP:conf/icassp/HanDLCZMX22}}} &\multirow{2}{*}{PS}& 4.88 & 5.54 & 5.78 & 5.80 & 5.29 & 5.59 & 5.80 & 5.45 \\
     && 94.42\textbar{}99.10\textbar{}96.70 & 85.68\textbar{}95.44\textbar{}90.30 & 82.10\textbar{}95.59\textbar{}88.34 & 82.71\textbar{}96.79\textbar{}89.20 & 90.00\textbar{}94.29\textbar{}92.09 & 87.68\textbar{}98.13\textbar{}92.61 & 83.36\textbar{}97.93\textbar{}90.06 & 61.91\textbar{}95.00\textbar{}74.97 \\
    \midrule
    \multirow{2}{*}{SeACo \cite{10446106}}&\multirow{2}{*}{ASF}&4.86 &4.89 & 4.80 & 4.76 & 5.24 & 5.10  & 5.01 & 5.06 \\
    &&93.99\textbar{}79.64\textbar{}86.22 &87.95\textbar{}95.44\textbar{}91.54 & 84.49\textbar{}96.61\textbar{}90.14 & 84.32\textbar{}97.59\textbar{}90.47 & 90.00\textbar{}69.72\textbar{}78.57 & 88.93\textbar{}96.38\textbar{}92.50 & 84.23\textbar{}97.43\textbar{}90.35 & 65.39\textbar{}91.37\textbar{}76.23 \\
    \midrule
    \multirow{2}{*}{PSC-Joint-P}&\multirow{2}{*}{GCP}&\textbf{4.74} &\textbf{4.62} & \textbf{4.49}  & \textbf{4.37} & \textbf{5.05} &  \textbf{4.81} &  \textbf{4.61} & \textbf{4.73} \\
    &&\textbf{97.00}\textbar{}97.84\textbar{}\textbf{97.41} &\textbf{92.02}\textbar{}95.63\textbar{}\textbf{93.79} & \textbf{90.32}\textbar{}96.18\textbar{}\textbf{93.16}  & \textbf{90.12}\textbar{}97.26\textbar{}\textbf{93.55} & \textbf{92.73}\textbar{}91.07\textbar{}\textbf{91.89}  &  \textbf{93.69}\textbar{}96.97\textbar{}\textbf{95.30} &  \textbf{90.90}\textbar{}97.26\textbar{}\textbf{93.97} & \textbf{80.67}\textbar{}94.35\textbar{}\textbf{86.98} \\
    \bottomrule
  \end{tabular}
\end{table*}

\subsection{Evaluation on Hyperparameter Sensitivity}
\begin{figure}[t!]
	\centering
  \includegraphics[height=4.15cm]{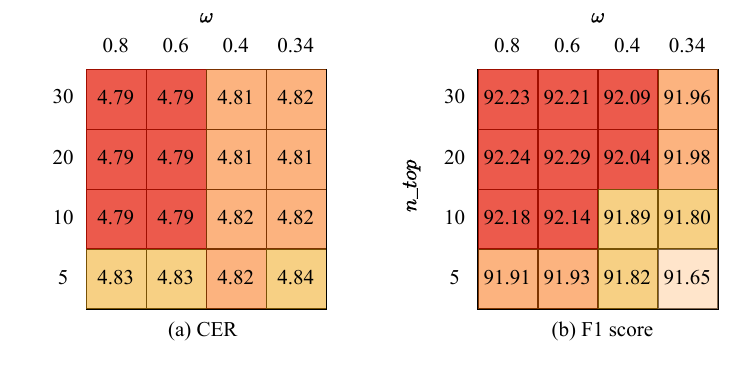}
  \caption{Sensitivity analysis of hyperparameter tuning for PSC-Joint-P (OCP) in terms of (a) CER (\%) and (b) F1 score (\%).}
  \label{fig:hyperOEM}
\end{figure}

To explore the impact caused by hyperparameter tuning, we analyze hyperparameter sensitivity for PSC-Joint-P on the AISHELL-1 task. 
The threshold for list-level correlation, $thres\_list$, is set to 0.5 without tuning, which is the median of the score interval $[0,1]$. The rest hyperparameters (the triangular window weight $\omega$, purification group size $group\_size$ , purification rounds $n\_r$, and the number of biasing phrases being kept $n\_top$) are split for the control variable study. In the PSC-Joint-P with the once competitive mechanism, $group\_size$ equals the length of the biasing list and $n\_r$ is 1, thus only $\omega$ and $n\_top$ need to be tuned.
Fig.~\ref{fig:hyperOEM} shows the tuned performance of the PSC-Joint-P (OCP) represented in the form of a heatmap, in which the $\omega$ is adjusted from 0.34 to 0.8 and the $n\_top$ is adjusted from 5 to 30. Overall, the fluctuations of CER and F1 score are minimal, indicating that the PSC-Joint-P (OCP) is insensitive to $\omega$ and $n\_top$. Thus, the $\omega$ is set to 0.6 and the $n\_top$ is set to 10 in our experiments.

\begin{figure}[t!]
	\centering
  \includegraphics[width=\linewidth]{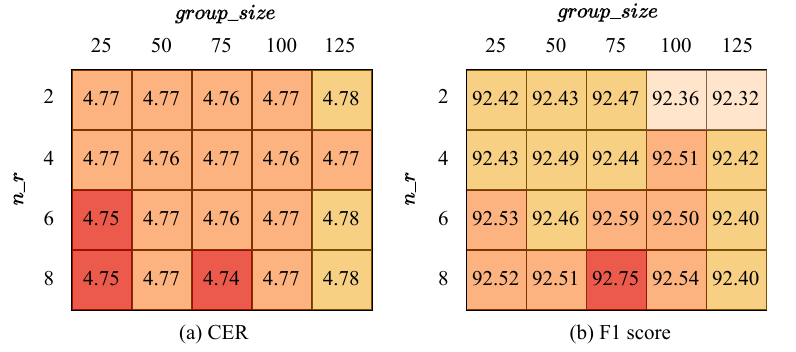}
  \caption{Sensitivity analysis of hyperparameter tuning for PSC-Joint-P (GCP) in terms of (a) CER (\%) and (b) F1 score (\%).}
  \label{fig:hyperGEM}
\end{figure}

With the determined $\omega$ and $n\_top$, we further explore the hyperparameter sensitivity of PSC-Joint-P (GCP) on $n\_r$ and $group\_size$. From Fig.~\ref{fig:hyperGEM}, it is clear that the PSC-Joint-P (GCP) model is insensitive to the rounds and group size. 
Considering the computational cost, the hyperparameter $n\_r$ is set to two. Another hyperparameter $group\_size$ is set to 75, closing to the average length of the biasing list during training. Notably, all experiments of PSC-Joint-P adopt the mentioned hyperparameter configuration, including the KeSpeech task.

\subsection{Evaluation on KeSpeech}
To explore the effect of the data scale, we conduct several experiments on the 895-hour data set, KeSpeech. In addition to the AISHELL-NER test set, the biasing lists of the AISHELL-NER development set are used as a supplement for evaluation, since it is not seen during the model training on KeSpeech. All model architectures, experimental settings, and hyperparameters are the same as the AISHELL-1 task. As shown in Table~\ref{tab:methodcomparitionkespeech}, all contextual ASR models achieve better performance in terms of F1 score, benefiting from the larger training data scale. Compared with ColDec, FineCoS, and SeACo, the PSC-Joint-P (GCP) model achieves 34.49\%, 33.40\%, and 43.76\% average relative F1 score improvements across varying-length biasing lists. For longer biasing lists, such as ``NE-2204'' (Dev set), the relative F1 score improvements of PSC-Joint-P (GCP) attach to 36.89\%, 40.28\%, and 32.32\% when compared to ColDec, FineCoS and SeACo, respectively. This demonstrates the strong robustness of PSC-Joint-P (GCP) in handling varying lengths of biasing lists. 

The ``PER-414'' biasing list has the highest bias OOV rate and lowest inclusion rate, which poses a significant challenge for the contextual ASR. PSC-Joint-P (GCP) outperforms the backbone with a relative improvement of 65.59\% on F1 score. For the issue of over-biasing, our PSC-Joint-P (GCP) attains a superior balance between the under- and over-biasing, i.e., lower CER and higher F1 score. This indicates that the most relevant biasing information is more accurately identified.

\begin{figure}[t!]
	\centering
	\includegraphics[height=13cm]{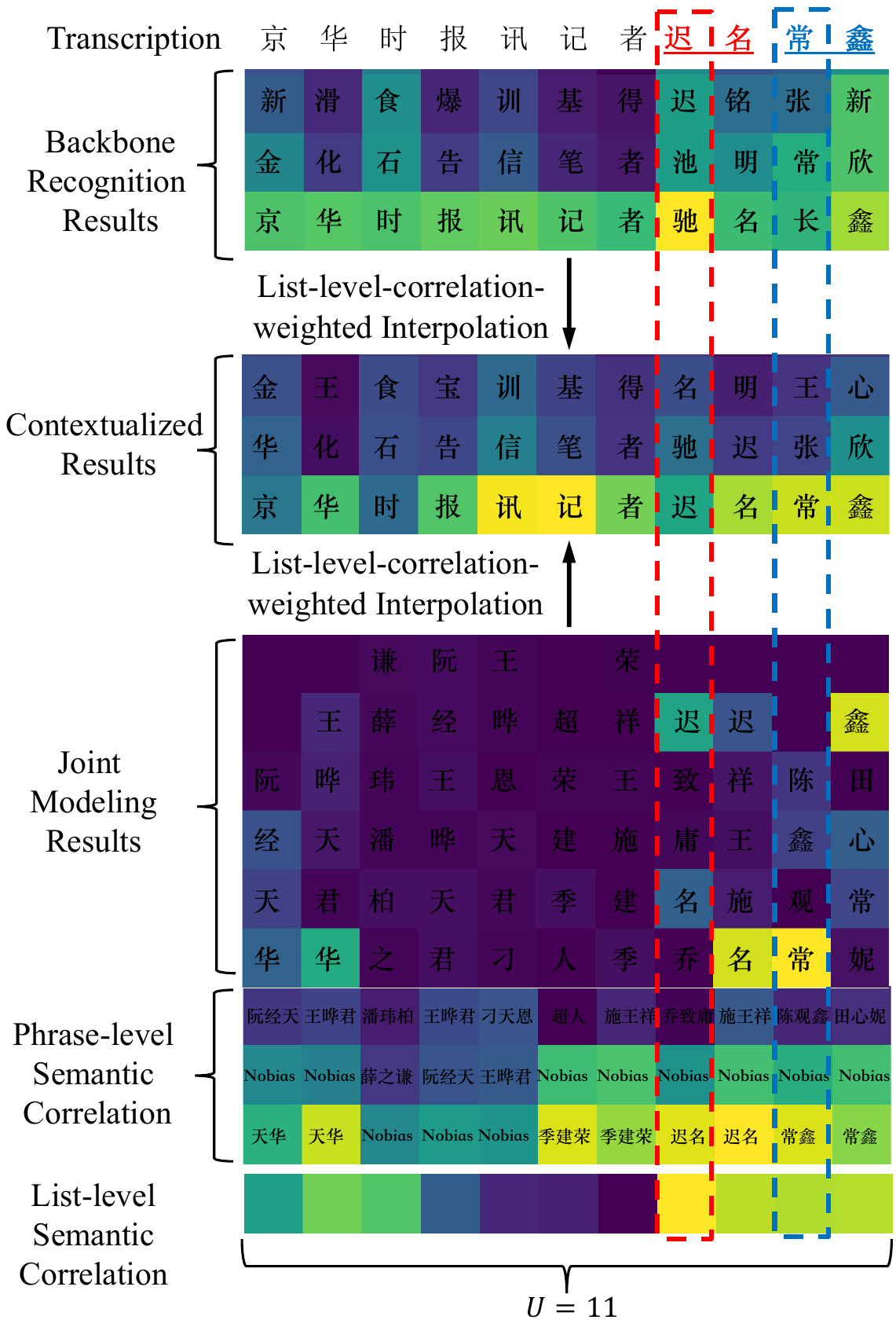}
	\caption{An instance. Note that, for demonstration, only key elements of the backbone recognition results, joint modeling results, and phrase-level correlation are illustrated, rather than the entire matrices. And lighter color indicates higher score. The red dashed box and the blue dashed box highlight the biasing process of tokens in two person names, respectively.}
	\label{fig:case}
\end{figure} 

\subsection{Instance Analysis}
To obtain a deeper understanding of the semantic correlation joint modeling, we analyze an instance by plotting the multi-level semantic correlations, the joint modeling results, the backbone recognition results, and contextualized results.
Fig.~\ref{fig:case} shows that the transcription has 11 tokens and two biasing phrases, \begin{CJK}{UTF8}{gbsn}\footnotesize ``迟名''\end{CJK} (Chi Ming) and \begin{CJK}{UTF8}{gbsn}\footnotesize ``常鑫''\end{CJK} (Chang Xin). The NAR model employs greedy search decoding, i.e., only the most relevant token will be output at each decoding step. However, \begin{CJK}{UTF8}{gbsn}\footnotesize ``迟''\end{CJK} (Chi) and \begin{CJK}{UTF8}{gbsn}\footnotesize ``常''\end{CJK} (Chang) are not the most relevant tokens at the seventh and ninth decoding steps in the backbone recognition results. Instead, \begin{CJK}{UTF8}{gbsn}\footnotesize ``驰''\end{CJK} (Chi) and \begin{CJK}{UTF8}{gbsn}\footnotesize ``长''\end{CJK} (Chang), which have the same pronunciation, are output. In the list-level correlation, the correlation score of indices from seven to ten are the highest, corresponding to the top-one biasing phrase \begin{CJK}{UTF8}{gbsn}\footnotesize ``迟名''\end{CJK} (Chi Ming), \begin{CJK}{UTF8}{gbsn}\footnotesize ``迟名''\end{CJK} (Chi Ming), \begin{CJK}{UTF8}{gbsn}\footnotesize ``常鑫''\end{CJK} (Chang Xin), and \begin{CJK}{UTF8}{gbsn}\footnotesize ``常鑫''\end{CJK} (Chang Xin) in the phrase-level correlation, respectively. After joint modeling, only a few tokens satisfy the cross-granularity consistency of the most relevant biasing information, achieving higher scores in the joint modeling results. Subsequently, the contextualized results are obtained by interpolating the backbone recognition and joint modeling results with the list-level correlation. In this manner, the right tokens in the seventh and ninth decoding steps are retrieved.



\section{CONCLUSION}
\noindent The cross-attention-based contextual ASR is sensitive to variations of biasing information volume. 
In this study, we propose the PSC-Joint approach to reduce the sensitivity by identifying and integrating the most relevant biasing information into the contextual ASR.
Specifically, we first define three semantic correlations at various granularities from coarse to fine: list-level, phrase-level, and token-level. Then, based on the cross-granularity consistency of the most relevant biasing information, the intersection of three semantic correlations is calculated through joint modeling. 
In addition, PSC-Joint purifies the irrelevant biasing phrases in a group and competitive manner to reduce the computational cost of the semantic correlation joint modeling.
Experiments evaluated on AISHELL-1, KeSpeech, and AISHELL-NER datasets demonstrate that PSC-Joint significantly improves the F1 score on varying-length biasing lists compared with baselines, enhancing the robustness of contextual ASR to variations in biasing information volume. Our PSC-Joint approach, combined with post-processing, also achieves a better trade-off between under- and over-biasing when introducing biasing information to ASR. Furthermore, the PSC-Joint approach speeds up the inference through biasing-phrase purification.

\section{Acknowledgments}
\noindent We sincerely thank Prof. Dong Wang (Tsinghua University) for valuable comments. This work was supported in part by National Natural Science Foundation of China under Grant No. 62376071.

\bibliographystyle{IEEEtran}
\bibliography{mybib}



\end{document}